\documentclass[%
 reprint,
 amsmath,amssymb,
 aps,
]{revtex4-2}

\usepackage[dvipdfmx]{graphicx}
\usepackage{dcolumn} 
\usepackage{bm}      
\usepackage{setspace}
\usepackage{amsfonts}
\usepackage{braket}

\newcommand{\del}[3] {\frac{\partial^{#3} #1}{\partial #2^{#3}}}
\newcommand{\dev}[3]{\frac{\text{d}^{#3} #1}{\text{d}#2^{#3}}}
\newcommand{\pdev}[3]{{\text{d}^{#3} #1}/{\text{d}#2^{#3}}}
\newcommand{\pdel}[3]{{\partial^{#3} #1}/{\partial #2^{#3}}}

\newcommand{\am}{{\bm a}}
\newcommand{\bb}{{\bm b}}

\newcommand{\pp}{{\bm p}}
\newcommand{\rr}{{\bm r}}

\newcommand{\tm}{{\bm t}}
\newcommand{\uu}{{\bm u}}

\newcommand{\ww}{{\bm w}}
\newcommand{\xx}{{\bm x}}
\newcommand{\yy}{{\bm y}}
\newcommand{\zz}{{\bm z}}

\newcommand{\RR}{{\mathbb R}}

\newcommand{\oomega}{\mbox{\boldmath $\omega$}}

\newcommand{\FF}{\mbox{\boldmath $F$}}

\newcommand{\GG}{\mbox{\boldmath $G$}}
\newcommand{\tr}{\mathrm{T}}

\usepackage{amsmath}	
\begin{document}

\preprint{APS/123-QED}

\title{
Physical deep learning based on optimal control of dynamical systems
}

\author{
Genki Furuhata$^1$, Tomoaki Niiyama$^2$, and Satoshi Sunada$^{2,3}$
}
 \email{sunada@se.kanazawa-u.ac.jp}
\affiliation{%
${}^1$Graduate School of Natural Science and Technology, Kanazawa University
Kakuma, Kanazawa, Ishikawa, 920-1192, Japan\\
${}^2$Faculty of Mechanical Engineering, Institute of Science and
Engineering, Kanazawa University
Kakuma-machi Kanazawa, Ishikawa 920-1192, Japan \\
${}^3$Japan Science and Technology Agency (JST), PRESTO, 4-1-8 Honcho,
 Kawaguchi, Saitama 332-0012, Japan\\
}%

\date{\today}

\begin{abstract}
Deep learning is the backbone of artificial intelligence technologies,
and it can be regarded as a kind of multilayer feedforward
 neural network. 
An essence of deep learning is information propagation through layers.
This suggests that there is a connection between deep neural
 networks and dynamical systems in the sense that 
information propagation is explicitly modeled by the time-evolution of
 dynamical systems. 
In this study, we perform pattern recognition based on the 
optimal control 
of continuous-time dynamical systems, which is suitable for 
physical hardware implementation. 
The learning is based on the adjoint method to optimally control 
dynamical systems, and the deep (virtual) network structures 
based on the time evolution of the systems 
are used for processing input information.
As a key example, we apply the dynamics-based recognition approach to 
an optoelectronic delay system 
and demonstrate that the use of the delay system allows for
image recognition and nonlinear classifications  
using {\it only a few} control signals.
This is in contrast to conventional multilayer 
neural networks, which require a 
large number of weight parameters to be trained.  
The proposed approach provides insight into the mechanisms of
 deep network processing in the framework of an optimal control problem
and presents a pathway for realizing physical computing hardware. 
\end{abstract}

\maketitle

\section{Introduction}
The recent rapid progress of information technologies, 
including machine learning,
has led to studies on novel computing concepts and hardware, 
such as neuromorphic processing \cite{Markovic2020,Rabinovich2006,Furber2016,Merolla2014,Feldmann2019,Tait2017}, 
reservoir computing \cite{Versraeten2007,Jaeger2004,Tanaka2019,Appeltant2011,Brunner2013,Inubushi2020}, 
and deep learning \cite{LeCun2015,Xu2018,Lin2018,Shen2017}.
In particular, deep learning has become a groundbreaking 
tool for data processing owing to its high-level performance \cite{LeCun2015}.
Furthermore, the energy-efficient computing for deep learning is gaining importance 
with the rising need for processing large amounts of data \cite{XChen2014}. 

An underlying key factor of deep learning is 
its high expressive power, which is the result of 
the layer-to-layer propagation 
of information in the deep network.
This expressive power enables the representation of 
extremely complex functions in a manner 
that cannot be achieved using shallow networks with the same number of neurons 
\cite{Poole2016,Montufar2014}. 
Interestingly, recent studies have reported
that the information propagation in 
multilayer systems can be expressed as the time evolution of 
dynamical systems \cite{Liu2019,Chen2018,Benning2019,Haber2017}. 
From the point of view of dynamical systems, the learning process of 
networks can be regarded as the optimal control of the dynamical systems 
\cite{Liu2019,Chen2018}. 
This viewpoint suggests that there is a 
connection between deep neural networks 
and dynamical systems and indicates the possibility of using
dynamical systems as physical deep-learning machines.

In this paper, we reveal the potential of dynamical systems 
with optimal control for the physical implementation.
We propose a deep neural network-like architecture 
using dynamical systems with delayed 
feedback and show that delayed feedback allows for
the virtual construction of a deep network structure 
in a {\it physically single node} using a time-division multiplexing
method.  
In the proposed approach, the virtual deep network for information propagation 
comes from the time evolution of delay systems; 
the systems are optimally controlled such that 
information processing, including classification, is facilitated.
The significant difference between our deep network 
and ordinary deep neural network architectures is 
that the learning via optimal control is realized 
by only a few control signals and minimal weight parameters,
whereas the learning by conventional deep neural networks requires
a large number of weight parameters \cite{Shen2017,Lin2018}.
The proposed approach using optimal control is applicable for 
a wide variety of experimentally controllable systems; 
it allows for simple but large-scale deep networks in physical systems 
with a few control parameters. 
\section{Multilayer neural networks and dynamical systems}
First, we briefly discuss the relationship 
between multilayer neural networks and dynamical systems. 
Let a dataset to be learned be composed of $K$ inputs, 
$\xx_{k} \in \RR^M$ and their corresponding target vectors, $\tm_k\in\RR^{L}$, 
where $k \in \{1,2,\cdots,K\}$. 
$M$ and $L$ are dimensions of the inputs and target vectors, respectively. 
The goal of supervised learning 
is to find a function that maps inputs 
onto corresponding targets, $\GG: \xx_k \rightarrow \tm_k$. 
To this end, we consider an output function,
$\yy = \tilde{\GG}(\xx,\ww) \in \RR^L$, 
parameterized by the $M_w$-dimensional vector, $\ww \in \RR^{M_w}$, 
and the following loss function:
\begin{align}
J = 
\sum_{k=1}^K
\Psi(\tm_k,\yy_k),  \label{cost-eq1}
\end{align}
where $\Psi(\tm_k,\yy_k)$ is a function of the distance between the target 
$\tm_k$ and the output, $\yy_k = \tilde{\GG}(\xx_k,\ww)$. 
$\ww$ is determined such that loss function $J$ is minimized, i.e., 
output $\yy_k$ corresponds to target $\tm_k$.

\begin{figure}[htbp]
\centering\includegraphics[width=7.5cm]{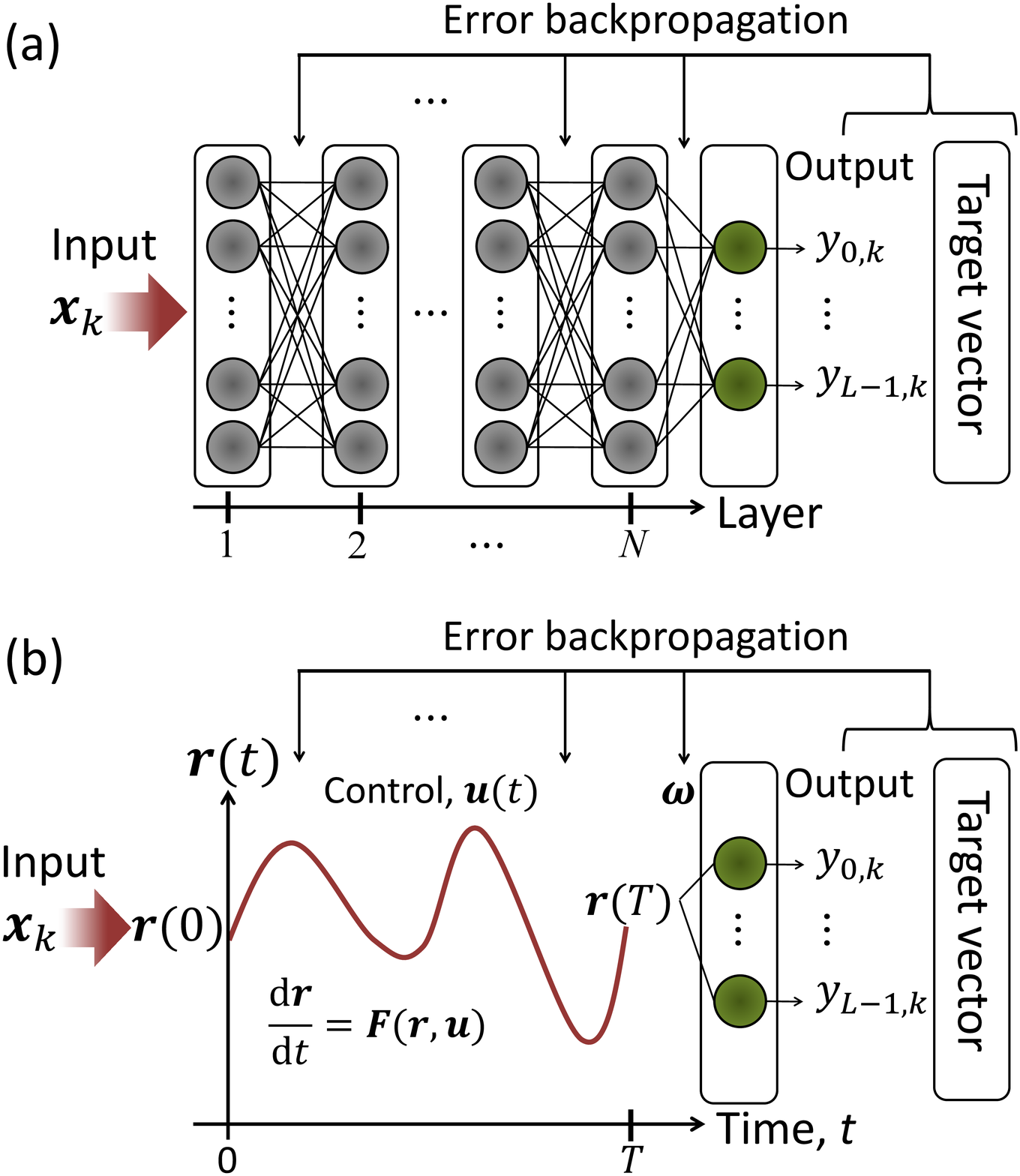}
\caption{\label{fig_ds}
Schematics of (a) multilayer neural network and (b) dynamical system.
In (b), the output, $\yy_k = (y_{0,k},y_{1,k},\cdots,y_{L-1,k})^\tr$, 
is given by the end state $\rr(T)$ and weight parameter $\oomega$. 
$\uu(t)$ and $\oomega$ can be updated 
using a gradient-based optimization algorithm.
}
\end{figure}

It is well-known that a neural network model with an appropriate 
activation function is a good candidate 
for representing function $\GG$ owing to its universal approximation 
capability \cite{Cybenko1989,Funahashi1989,Sonoda2017}. 
In multilayer neural networks, 
the output, $\yy_k = (y_{0,k},y_{1,k},\cdots,y_{L-1,k})^\tr$, 
is given by the layer-to-layer propagation of an input, $\xx_k$
[Fig.~\ref{fig_ds}(a)]. 
The layer-to-layer propagation based on multilayer network structures 
plays a crucial role in increasing expressivity \cite{Poole2016} 
and enhancing learning performance. 
In this study, instead of standard multilayer networks,  
we utilize information propagation in 
a continuous-time dynamical system,
\begin{align}
\dev{\rr(t)}{t}{} = \FF\left[\rr(t),\uu(t)
\right], \label{eq1}
\end{align}
where $\rr(t) \in \RR^M$ is the state vector at time $t$
and $\uu(t) \in \RR^{M_u}$ represents a control signal vector. 
Based on the correspondence between a multilayer network [Fig.~\ref{fig_ds}(a)] 
and a dynamical system [Fig.~\ref{fig_ds}(b)],
we suppose that 
an input $\xx_k$ is set as an initial state $\rr(0)$.
Additionally, the corresponding output, $\yy_k$, 
is given by the time evolution 
(feedforward propagation) of 
the state vector, $\rr_k(T) = \rr(T,\xx_k)$, up to the end time $t = T$, 
i.e., $\yy_k = \yy[\rr_k(T),\oomega]$, where 
$\oomega \in \RR^{L\times M}$ is a parameter matrix determined in 
the training process. 
Loss function $J$ is obtained by repeating the aforementioned feedforward 
propagation for all training data instances and using their outputs.  
The goal of the learning is to find 
an optimal control vector, $\uu^*(t)$, and a parameter vector, 
$\oomega^*$, such that $J$ is minimized, i.e.,  
$\ww^* = (\{\uu^*(t)\}_{0 < t \le T}, \oomega^*) = \mbox{argmin}_{\ww} J$.
It should be noted that learning using a discretized version of Eq. (\ref{eq1}) directly 
corresponds to that using a residual network (ResNet) \cite{Chen2018,Benning2019}.

One strategy for finding optimal controls and parameters is to compute
the gradients of the loss function, i.e., the direction of the steepest
descent, and to update control and parameter vectors
in an iterative manner, i.e, 
$\uu(t) \rightarrow \uu(t) + \delta \uu(t)$ and 
$\oomega \rightarrow \oomega + \delta \oomega$, respectively.  
In a simple gradient descent method, 
$\delta \uu(t)$ and $\delta \oomega$ are chosen such that 
the largest local decrease of $J$ is obtained.  
$\delta \oomega$ is simply selected
in the opposite direction of the gradient $\pdev{J}{\oomega}{}$, 
e.g., $\delta \oomega = -\alpha_{\omega}\pdev{J}{\oomega}{}
=
-\alpha_{\omega}\sum_k(\pdel{\Psi}{\yy_k}{}\pdel{\yy_k}{\oomega}{})$, 
where $\alpha_{\omega}$ is the learning rate, 
which is usually a small positive number.   
$\delta \uu(t)$ is obtained using the adjoint method 
developed in the context of optimal control problems \cite{Kirk2004,Sage1977} 
as follows:
\begin{align}
\delta\uu(t)
=
-\alpha_{u}
\sum_{k=1}^K
\left(
\pp_{k}^\tr(t)\del{\FF_{k}}{\uu}{}
\right)^\tr,
\label{eq1-j}
\end{align}
where $\alpha_u$ is a small positive number, 
$\FF_k = \FF[\rr_k(t),\uu(t)]$ and $\rr_k(t) = \rr(t,\xx_k)$. 
$\pp_{k}(t) \in \RR^M$ is the adjoint state vector that satisfies 
the end condition at $t = T$, 
$\pp_{k}(T) = \pdel{\Psi[\tm_k,\yy_k(\oomega,\rr_k)]}{\rr_k}{}|_{t = T}$. 
For $0 \le t < T$, the time evolution of $\pp_{k}(t)$ is given 
by 
\begin{align}
\dev{\pp_{k}^\tr(t)}{t}{}
= - \pp_{k}^\tr (t)\del{\FF_{k}}{\rr_{k}}{}.
\label{eq1-p}
\end{align}
The derivation of Eqs. (\ref{eq1-j}) and
(\ref{eq1-p}) is shown in Appendix \ref{app1}. 
We note that integrating Eq. (\ref{eq1-p}) in the backward direction (from $t=T$ to
$t=0$) corresponds to backpropagation in neural networks. 
In summary, the algorithm for computing optimal $\uu^*$ and $\oomega^*$ is 
as follows:\\
\noindent (i) Set the input, $\xx_k$, for the $k$th data instance as 
an initial state, i.e., $\rr(0) = \xx_k$.\\
\noindent (ii) Forward propagation: 
Starting from initial state $\xx_k$, integrate Eq. (\ref{eq1}) 
and obtain the end state, $\rr_k(T) = \rr(T,\xx_k)$. 
Then, compute the output, $\yy_k = \yy[\rr_k(T),\oomega]$. \\
\noindent (iii) Repeat the forward propagation for all data instances 
and compute loss function $J$. \\ 
\noindent (iv) Backpropagation: Integrate adjoint Eq. (\ref{eq1-p})
for $\pp_{k}(t)$ in the backward direction from $t = T$ 
with $\pp_{k}(T) = \pdel{\Psi}{\rr_{k}(T)}{}$. \\
\noindent (v) Compute $\delta\oomega$ 
and $\delta\uu(t)|_{0 < t < T}$ using  Eq. (\ref{eq1-j}) with
appropriate learning rates, $\alpha_{\omega}$ and $\alpha_{u}$.\\ 
\noindent (vi) Update control signal $\uu(t)$ and parameter $\oomega$.
For the updates, one can use different optimization algorithms 
\cite{Ruder2016}.   

\subsection{Binary classification problem \label{sec_ode}}
\begin{figure}[htbp]
\centering\includegraphics[width=8.8cm]{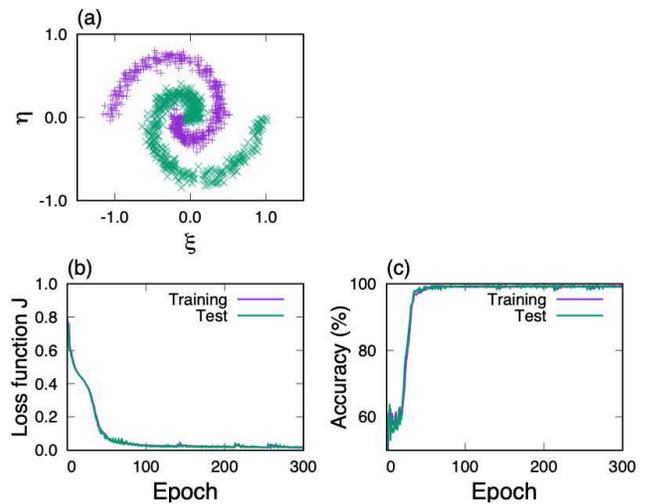}
\caption{\label{fig_ode1}
(a) Spiral dataset for binary classification. 
The dataset consists of the two data groups labeled as ``0'' or ``1'', 
colored purple or green, respectively. 
(b) Loss function $J$ and (c) classification accuracy as a function of 
the training (test) epoch. 
}
\end{figure}

Here, we use an abstract dynamical system 
for solving a typical fundamental problem, the binary classification problem. 
The goal of the binary classification is to classify a given dataset 
into two categories labeled as, for example, ``0'' or ``1''. 
For this, we here consider a simple dynamical model, 
$\dot{\rr} = \tanh[\am(t)\rr+\bb(t)]$, where
the state vector is two-dimensional, $\rr =(\xi,\eta)^\tr$. 
Weight $\am(t) \in \RR^{2\times 2}$ and 
bias $\bb(t) \in \RR^2$ are used as control signals.
We apply this model to the binary classification problem for 
a spiral dataset, $\{\xx_k,c_k\}_{k=1}^K$, where $\xx_k \in \RR^2$ 
is distributed around 
one of two spirals in the $\xi\eta$-plane, as shown in 
Fig.~\ref{fig_ode1}(a), and $\xx_k$ 
is labeled by $c_k$ as ``0'' or ``1'' according to the classes. 
For the classification, we used one-hot encoding, i.e.,  
target $\tm_k$ corresponding to input $\xx_k$ was set 
as $\tm_k = (t_{0,k},t_{1,k})^\tr = (1,0)^\tr$ if $c_k = 0$ 
and $\tm_k = (0,1)^\tr$ if $c_k = 1$.
For the output, $\yy_k = (y_{0,k},y_{1,k})^\tr$, the softmax function
was used: $y_{l,k}= \exp(z_{l,k})/\sum_l\exp(z_{l,k})$, where 
$z_{l,k} = \oomega_l^\tr\rr_k(T) + \omega^{bias}_{l}$,  
$\oomega_l \in \RR^2$, and $\omega^{bias}_{l} \in \RR$.
If $z_{0,k} \gg z_{1,k}$, $\yy_k$ approaches $(1,0)^\tr$, 
whereas if $z_{0,k} \ll z_{1,k}$, $\yy_k$ approaches $(0,1)^\tr$.
$J$ was selected as a cross-entropy loss function, 
$J =  -1/K\sum_{k=1}^K \sum_{l=0}^1t_{l,k}\ln y_{l,k}$. 
A training set of $K = 1000$ data points was used to 
train $\uu(t)=\{\am(t),\bb(t)\}$ and 
$\oomega = \{\oomega_l,\omega^{bias}_{l}\}_{l=0,1}$.
The classification accuracy was evaluated 
using a test set of $1000$ data points.   
For the gradient-based optimization, we used the Adam optimizer 
\cite{Kingma2014} with a batch size of $K$. 

\begin{figure}[htbp]
\centering\includegraphics[width=8.8cm]{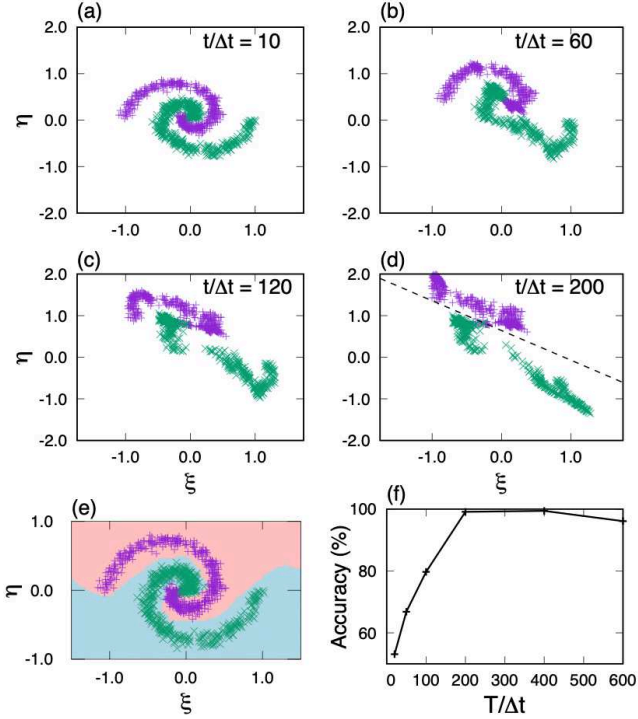}
\caption{\label{fig_ode2}
(a--d) Configuration of the states constituting the two spirals 
over the time evolution in the trained system 
up to the end time $T=200\Delta t$.
$t/\Delta t$ effectively represents the number of layers from the viewpoint of 
a neural network.   
The initial spiral distribution is disentangled according to
the time evolution (layer-to-layer propagation) and 
becomes linearly separable at end time $t = T$.
In (d), the dotted line represents the decision boundary for separating 
the two distributions.
(e)  
Result of binary classification.
The inputs can be classified into two regions 
indicated by pink and blue colors. 
(f)   
Classification accuracy for a test dataset 
as a function of end time $T$.
}
\end{figure}

Figures \ref{fig_ode1}(b) and \ref{fig_ode1}(c) show 
the learning curve and classification accuracy
for the training and test datasets, respectively.
The loss function monotonically decreases, and the 
classification accuracy approaches 100 $\%$.  
For sufficient training over 300 training epochs, 
the classification accuracy was over 99 $\%$ 
when end time $T$ is set as $200\Delta t$, 
where $\Delta t \approx 0.01$ is the time step used in the simulation.
Figure \ref{fig_ode2} shows the time evolution of the two distributions
constituting the spiral dataset.
During the evolution, 
the distributions of the initial states are disentangled 
[Figs.~\ref{fig_ode2}(a-d)] and become 
linearly separable at end time $T$ [Fig.~\ref{fig_ode2}(d)]
to aid the classification at the softmax output layer. 
As a result, any input state can be classified into either of 
two classes [Fig.~\ref{fig_ode2}(e)].
We note that 
the classification based on disentanglement 
is different from 
that of other schemes utilizing dynamical systems, e.g., 
reservoir computing, 
whose classification is based on the mapping of 
input information onto a high-dimensional feature space \cite{Tanaka2019}.  
In addition, we note that the disentanglement is facilitated as end time 
$T$ increases, i.e., the number of layers increases,   
and high classification accuracy is achieved for $T  \le 400\Delta t$, 
as shown in Fig. \ref{fig_ode2}(f).  
A slight decrease in classification accuracy at $T = 600 \Delta t$ 
is attributed to slowdown of the training due to 
a local plateau of loss function $J$,
which is occasionally caused in a non-convex optimization problem
\cite{Dauphin2014}. 
At $T =600\Delta t$, 
we confirmed that 
a classification accuracy of over 99 $\%$ was obtained
when the number of training epochs is extended to 400.

\section{Physical implementation in delay systems}
In the previous section, we showed binary classification 
based on optimal control in a two-dimensional dynamical model, 
where weight $\am(t) \in \RR^{2\times 2}$ and bias $\bb(t) \in \RR^2$ 
were used as control signals.  
The excellent performance of this demonstration suggests
the realization of deep learning in various physical systems, 
such as coupled oscillators, fluids, and elastic bodies.
However, for processing high-dimensional data,
a number of signals must be used to control 
the high-dimensional degrees of freedom in the systems;
this may be difficult 
in terms of physical implementation in an actual system.  

To overcome the difficulty, we propose the use of delay systems 
to achieve feasible optimal control of numerous degrees of freedom
with limited control signals.
It is known that delay systems can be regarded as 
infinite-dimensional dynamical systems, as visualized  
in a time--space representation \cite{Arecchi1992}.
Furthermore, delay systems can support numerous virtual neurons 
using a time-division multiplexing method \cite{Appeltant2011}. 
In addition, they can exhibit 
various dynamical phenomena, including stable motion, 
periodic motion, and high dimensional chaos, with experimentally controllable
parameters, e.g., delay time and feedback strength 
\cite{Uchida,Soriano2013}.
Thus, their high expressivity as well as controllability 
are promising. 

\subsection{Learning by optimal control in a delay system \label{sec_delay}}
We introduce a training method based on 
the optimal control of a delay system, 
the time evolution of which 
is governed by the following equation:  
\begin{align}
\dev{\rr(t)}{t}{} 
= \FF\left[\rr(t),\rr({t-\tau}),\uu(t)
\right], \label{delay-eq}
\end{align}
where $\rr(t) \in \RR^{M_r}$ and $\uu(t) \in \RR^{M_u}$ 
represent the state vector and control signal 
vector at time $t$, respectively, and $\tau$ is the delay time.  
The aforementioned equation can be integrated by setting
$\rr(t)$ for $-\tau \le t \le 0$ as an initial condition.  

\begin{figure}[htbp]
\centering\includegraphics[width=8cm]{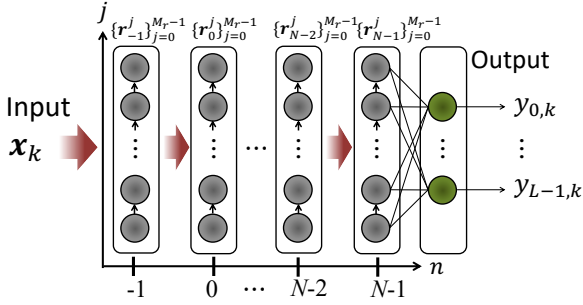}
\caption{\label{fig_dn}
Schematic of virtual network of a delay system.}
\end{figure}

The information dynamics can intuitively be interpreted by 
a space--time representation \cite{Arecchi1992} 
based on the time discretization of Eq. (\ref{delay-eq}), 
$\rr_{n}^{j+1} = \rr_n^j +\Delta t\FF(\rr_n^j,\rr_{n-1}^j,\uu_n^j)$, 
where $t = n\tau + j\Delta t$, $\rr_n^j = \rr(n\tau + j\Delta t)$, 
$n \in \{-1,0,1,\cdots, N-1\}$, $j \in \{0,1,\cdots, M_{\tau}-1\}$, 
and $M_{\tau} = \tau/\Delta t$. 
In this representation, $\rr_n^j$ can be regarded as 
the $j$th network node in the $n$th layer,
which is affected by an adjacent node, $\rr_n^{j-1}$, 
and node $\rr_{n-1}^j$ in the $(n-1)$th layer, as shown in 
Fig. \ref{fig_dn}. 

Feedforward propagation is carried out as follows:
First, an input, $\xx_{k} = (x_{1,k},\cdots,x_{M,k})^\tr$,
is encoded as $\rr_{-1}^{j} = \rr_{-1}^j(\xx_{k})$ 
for $j \in \{0,1,\cdots, M_{\tau}\}$
in the initial condition.  
Then, Eq. (\ref{delay-eq}) is numerically solved to obtain 
$\rr_{N-1}^j$ in the $(N-1)$th layer (corresponding to 
$\{\rr_k(t)\}_{T-\tau \le t < T}$ in continuous time).
The output, $\yy_k \in \RR^L$, is computed using the nodes 
in the $(N-1)$th layer, $\{\rr_{N-1}^j\}_{j=0}^{M_\tau-1}$, 
as shown in Fig. \ref{fig_dn}. 
In the continuous time representation, the output is defined as 
$\yy_k = 
\yy\left(
\zz_{k}
\right)
$,  
where 
$\zz_{k} = \int_{T-\tau}^T\oomega(t)
\rr_{k}(t)dt + \bb$.
$\oomega(t) \in \RR^{L\times M_{r}}$ and $\bb \in \RR^L$
are the weight and bias parameters to be trained, respectively, 
and $\rr_k(t)$
is the state vector starting from the initial state $\{\rr(t,\xx_k)\}_{-\tau \le t \le 0}$. 
Finally, loss function $J$ [Eq. (\ref{cost-eq1})]
is computed.

To minimize loss function $J$, a gradient-based optimization is used, 
where $\uu(t)$, $\oomega(t)$, $\bb$ are updated in an iterative manner.
In a gradient descent method, the 
update variations, $\delta\uu(t)$, $\delta\oomega(t)$, and $\delta\bb$, 
can be chosen as follows:
\begin{align}
\delta\uu(t)
= -\alpha_{u}\sum_{k=1}^K
\left(
\pp_{k}^\tr(t)\del{\FF_{k}}{\uu}{}
\right)^T, \label{update-delayeq1}
\end{align}
\begin{align}
\delta\oomega
= -\alpha_{\omega}
\sum_{k=1}^K
\left(
\rr_k
\del{\Psi}{\zz_{k}}{}
\right)^\tr,
\hspace{1mm}
\delta\bb_l 
= -\alpha_{b}
\sum_{k=1}^K
\left(
\del{\Psi}{\zz_{k}}{}
\right)^\tr. 
\end{align}
The details of these derivations are provided in Appendix~\ref{app2}. 
In the aforementioned equations, 
$\FF_{k} = \FF[\rr_{k}(t),\rr_{k}(t-\tau), \uu(t)]$, 
$\Psi = \Psi(\tm_k,\yy(\zz_k))$ is a function of $\zz_k$, 
and 
$\alpha_{i}$ for $i \in \{u,\omega,b\}$ is the learning rate 
which is a small positive number.
In Eq. (\ref{update-delayeq1}), 
$\pp_{k}(t)$ is the adjoint state vector, 
which satisfies $\pp_k(T)=0$.
$\pp_k(t)$ can be obtained by solving the 
following adjoint equations in 
the backward direction, 
\begin{align}
\del{\pp_{k}^\tr (t)}{t}{}
= 
-\del{\Psi}{\zz_{k}}{}\oomega(t)
-
\pp_{k}^{\tr}(t)
\dfrac{\partial \FF_{k}}{\partial \rr_{k}}, \label{delay-eqp1} 
\end{align}
for $T-\tau \le t < T$, and
\begin{align}
\dev{\pp_{k}^\tr (t)}{t}{}
=
-\pp^\tr (t)
\dfrac{\partial \FF_{k}}{\partial \rr_{k}}
-\pp_{k}^\tr (t+\tau)
\dfrac{\partial \FF_{k}(t+\tau)}{\partial \rr_{k}}, \label{delay-eqp2} 
\end{align}
for $0 \le t < T-\tau$. 

\subsection{Optoelectronic delay system}
\begin{figure}[htbp]
\centering\includegraphics[width=8cm]{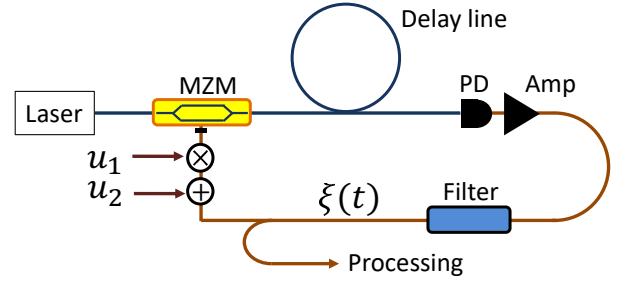}
\caption{\label{fig_delaysys}
Schematic of optoelectronic delay system. MZM, Mach--Zehnder modulator;
 Delay line, optical fiber delay line; PD, photodetector; Amp, electric
 amplifier; Filter, a two-pole band-pass filter (consisting of low-pass and high-pass filters).
}
\end{figure}

As an effective and feasible example, 
we consider the use of an optoelectronic delay system, 
as shown in Fig.~\ref{fig_delaysys}. 
The delay system is
composed of a laser, optoelectronic intensity modulator, photodetector,
and electrical filter to construct a time-delay feedback loop. 
The time evolution of the system state, 
$\rr(t) = (\xi(t),\eta(t))^\tr$, is given
by the following equations \cite{Murphy2010}:
\begin{align}
\tau_L\dfrac{d\xi}{dt} = 
-\left(
1 + \dfrac{\tau_L}{\tau_H}
\right)\xi - \eta 
+ \beta \cos^2
\left[
u_1(t)\xi(t-\tau)
+u_2(t)
\right], \label{eq-optdelay1}
\end{align}
\begin{align}
\tau_H\dfrac{d\eta}{dt} = \xi, \label{eq-optdelay2}
\end{align}
where
$\xi(t)$ is the normalized voltage, and 
$\tau_H$ and $\tau_L$ are the time constants of the low-pass and
high-pass filters, respectively. 
$\beta$ represents the feedback strength.
$u_1(t)$ and $u_2(t)$ are electronic signals added to the feedback loop, 
which are used as control signals in the system. 

\subsection{Results}
\subsubsection{Binary classification}
We demonstrate binary classification for a spiral dataset, as shown in 
Fig.~\ref{fig_ode1}(a), 
with the aforementioned optoelectronic delay system. 
The goal of binary classification is to classify two categories labeled 
as ``0'' or ``1'' for the spiral dataset, $\{\xx_k, c_k\}_{k=1}^K$.  
In the same manner as that described in Sec. \ref{sec_ode},
target $\tm_k$ is set 
as $\tm_k = (1,0)^\tr$ if $c_k = 0$, 
whereas $\tm_k = (0,1)^\tr$ if $c_k =1$. 
Output $y_{l,k}$ is set as the softmax function, i.e., 
$y_{l,k} = \exp{z_{l,k}}/\sum_{l^{'}}\exp{z_{l^{'},k}}$,
where 
$z_{l,k} = \int_{T-\tau}^T \omega_l(t)\xi_k(t) dt + b_l$, 
$\omega_l(t) \in \RR$ and $b_l \in \RR$.
Then, $J$ is defined as 
the cross-entropy loss function, $-1/K\sum_{k=1}^K\sum_{l=0}^{L-1}t_{l,k}\log{y_{l,k}}$. 
In this simulation, the following parameters settings were applied: 
$\tau_H = 1.59$ ms, $\tau_L = 15.9$ $\mu$s, and $\tau = 230$ $\mu$s. 
The input, $\xx_k = (x_{1,k},x_{2,k})^\tr$, 
is encoded as the initial state of $\xi$, i.e., 
$\xi_k(t) = x_{1,k}$ for $-\tau \le t < -\tau/2$ and 
$\xi_k(t) = x_{2,k}$ for $-\tau/2 \le t \le 0$.
We set $u_1(t) = 1.0$ and $u_2(t) = -\pi/4$ as the initial control signals
and $\omega_l(t) = 0$ and $b_l = 0$ ($l \in \{0, 1\}$) 
as the initial weight and bias parameters. 
We used the Adam optimizer \cite{Kingma2014} with a batch size of $K$
for the gradient-based optimization.  
The update equations for $u_1(t)$, $u_2(t)$, $\omega_l$, 
and $b_l$ are shown in Appendix~\ref{sec_app2-2}. 

\begin{figure}[htbp]
\centering\includegraphics[width=8.8cm]{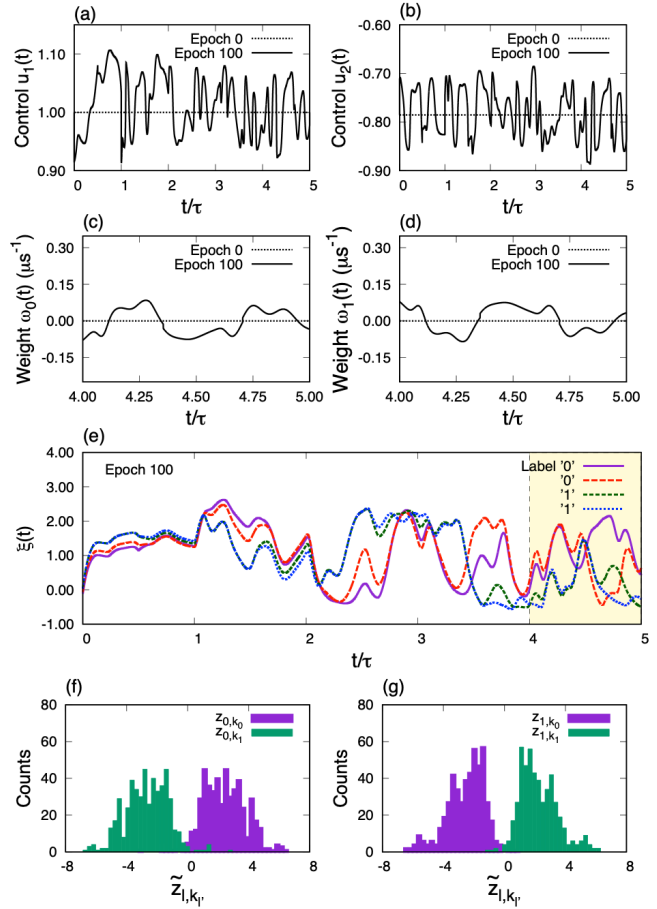}
\caption{\label{fig_dbr1}
(a,b) Control signals, $u_1(t)$ and $u_2(t)$ 
at training epoch 100.
(c,d) Weight parameters, $\omega_0(t)$ and $\omega_1(t)$, 
at training epoch 100.
(e) Four instances of $\xi_k(t)$ starting from different initial states
labeled as ``0'' or ``1'', which are 
within a distance $|\xx_k-\xx_{k^{'}}| < 0.1$.
The end time is set as $T = 5\tau$. 
$\xi_k(t)$ for $4\tau \le t < 5\tau$ (indicated by light yellow color)
is used to obtain $z_{l,k} = \int_{T-\tau}^T\omega_l(t)\xi_k(t)dt + b_l$.  
(f,g) Histograms of correlation values, $\tilde{z}_{l,k_{l'}}$, 
between softmax weights, $\omega_l(t)$, and the $k_{l'}$-th 
instance, $\xi_{k_{l'}}$. 
}
\end{figure}

In the aforementioned conditions, 
the classification accuracy at training epoch 100 was 99.1$\%$ 
when the feedback strength was $\beta = 3.0$ and the end time was 
$T = 5\tau$.
To gain insight into the classification mechanism, we 
investigated the effect of the control signals, $u_1(t)$ and $u_2(t)$, 
and weights, $\omega_l(t)$, $l \in \{0,1\}$, on the delay dynamics. 
Figures~\ref{fig_dbr1}(a--e) show the trained control signals, $u_1(t)$ and $u_2(t)$, 
weights, $\omega_0(t)$ and $\omega_1(t)$, 
and four instances of $\xi_k(t)$ at training epoch 100. 
$\xi_k(t)$ in a range of $T-\tau \le t \le T$ 
is used for computing the 
(softmax) outputs, $y_{l,k}$, which 
represents the probability that the input $\xx_k$ is 
classified as class $l \in \{0,1\}$.
Considering that $y_{l,k}$ is a function of
$z_{l,k} = \int_{T-\tau}^T\omega_l\xi_k(t)dt + b_l$, 
we computed $\tilde{z}_{l,k_{l'}} 
= \int_{T-\tau}^T\omega_l(t)\xi_{k_{l'}}(t) dt$.
$\tilde{z}_{l,k_{l'}}$ corresponds to the correlation between 
(softmax) weights $\omega_l(t)$ used for the classification 
as class $l$ and the $k_{l'}$th instance, $\xi_{k_{l'}}(t)$, which starts 
from the initial states labeled as $l' \in \{0,1\}$. 
Figures~\ref{fig_dbr1}(f) and \ref{fig_dbr1}(g) show 
the histograms of the correlation values, $\tilde{z}_{l,k_{l'}}$, 
for 500 instances.
The correlation values are positive for $l = l'$ in most cases, 
whereas they are negative for $l \ne l'$.
In other words, $\tilde{z}_{l,k_{l}} > \tilde{z}_{l,k_{l'}} (l \ne l')$ 
in most cases. 
Thus, the softmax output $y_{l,k_{l}}$ for classification as $l$ is 
activated by $\xi_{k_l}(t)$ starting from 
the initial states with the same class $l$.
These results reveal that 
the trained control signals control 
each trajectory such that
it is positively correlated to the weight $\omega_l(t)$ 
and $y_{l,k_l}$ is maximized. 

\begin{figure}[htbp]
\centering\includegraphics[width=8.8cm]{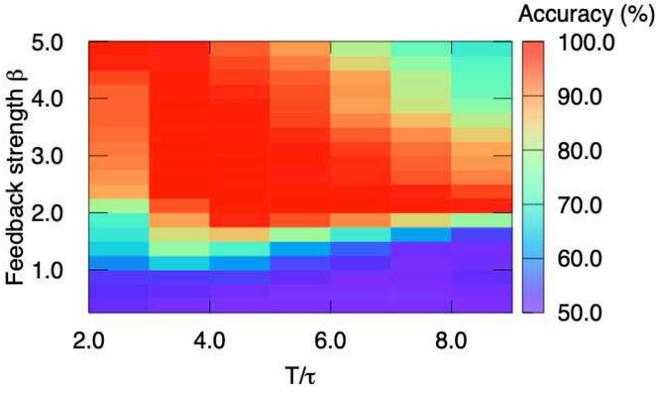}
\caption{\label{fig_dbr2}
Classification accuracy as a function of feedback strength $\beta$ 
and end time $T$. 
}
\end{figure}

\begin{figure}[htbp]
\centering\includegraphics[width=8.8cm]{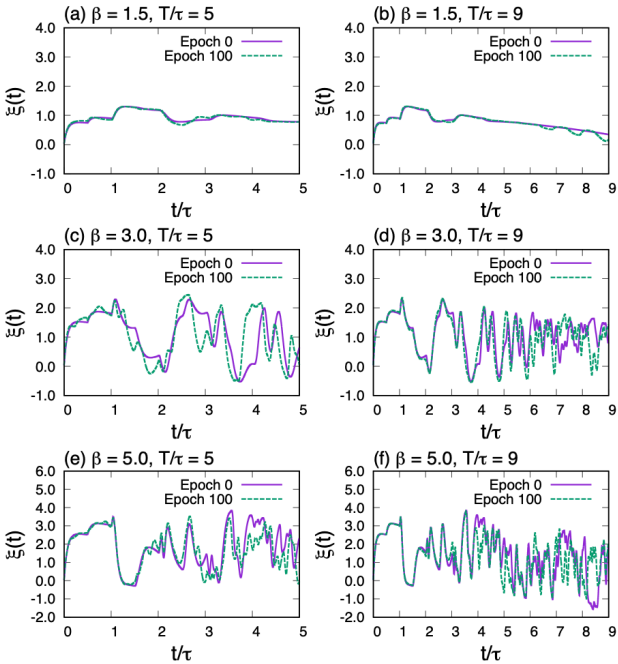}
\caption{\label{fig_dbr3}
Typical examples of $\xi_k(t)$ starting from an initial state  
at training epochs 0 and 100 for various values of $\beta$ and $T$. 
}
\end{figure}

Figure \ref{fig_dbr2} shows the classification accuracy at training epoch 100 
as a function of feedback strength $\beta$ and end time $T/\tau$.
As seen in this figure, 
classification performance is low for $\beta < 2.0$. 
In this regime, the system exhibits transient behavior 
to stable limit cycle motion, which 
is insensitive to external perturbations 
[Figs.~\ref{fig_dbr3}(a) and \ref{fig_dbr3}(b)].
This means that it is difficult to control the system.
When $\beta$ increases ($\beta > 2.5$), 
the system starts to exhibit complex behavior and 
becomes sensitive to the control signals for a large end time, $T$,
as shown in Figs. \ref{fig_dbr3}(c) - \ref{fig_dbr3}(f). 
Sensitivity plays a role in aiding the classification.
However, 
extremely high sensitivity makes it difficult to control the system 
and decreases classification performance,
as observed for $\beta > 4.0$ and $T > 7.0\tau$ in Fig.~\ref{fig_dbr2}.
Consequently, a high classification performance of over 99$\%$ is achieved
with moderate values of $T$ and $\beta$, suggesting 
that the transient behavior around the edge of chaos plays a crucial role in 
classification.  

\subsubsection{MNIST handwritten digit classification}
To investigate the classification performance for a higher dimensional dataset, 
we use the MNIST handwritten digit dataset, commonly  
used as a standard benchmark for learning \cite{MNIST,LeCun1998}.
The dataset has a training set of 60,000 28$\times$28 pixel grayscale images of ten handwritten digits, along with a test set of 10,000 images.

\begin{figure}[htbp]
\centering\includegraphics[width=8.8cm]{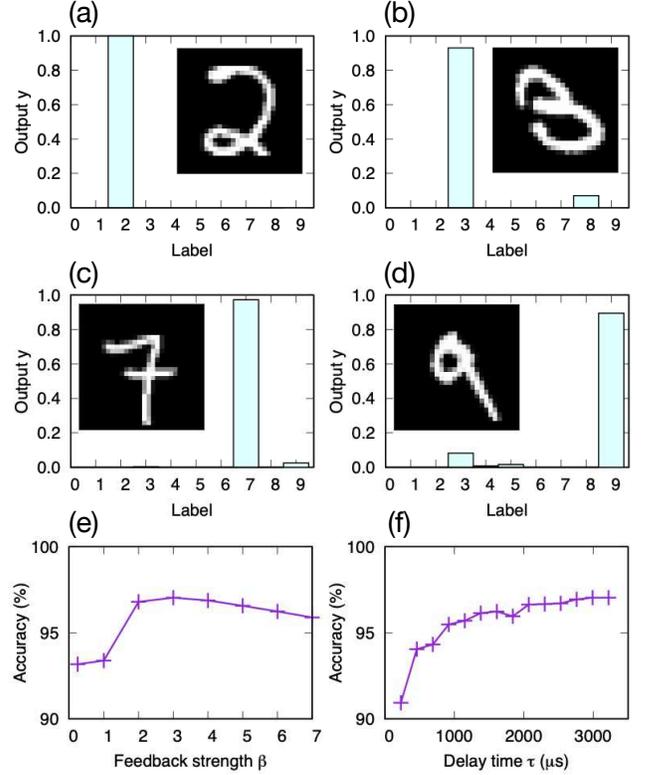}
\caption{\label{fig_dmr}
(a-d) Examples of softmax outputs, $\{y_{l,k}\}_{l=0}^9$, 
for the MNIST handwritten images. 
Each inset shows the input handwritten image corresponding to 
(a) ``$2$'', (b) ``$3$'', (c) ``$7$'', and (d) ``$9$''. 
The softmax output, $y_{l,k}$, represents the probability that 
the input image $k$ belongs to class $l$.  
In (a--d), $\beta = 3.0$, $\tau = 1610$ $\mu$s, and $T = 3\tau$.
(e--f) Classification accuracy for the MNIST dataset (10,000 test images) 
as a function of 
(e) feedback strength $\beta$ and (f) delay time $\tau$ 
in the delay system.
In (e), $\tau = 3220$ $\mu$s and $T = 3\tau$. 
In (f), $\beta = 3.0$ and $T = 3\tau$. 
}
\end{figure}

To set an initial state of the system state,
an input image of $28\times 28$ pixels is enlarged to double its length
and width; it is transformed to a $m$-dimensional vector, where 
$m = (28\times 2)^2$. 
The vector components are sequentially set 
as $\xi_k(t_j)$ at time $t_j = -\tau + j\Delta t$ 
with a time interval of $\Delta t = \tau/M_{\tau} \approx 0.07$ $\mu$s.
The input process is repeated $M_{\tau}/m$-times 
to encode the information of the input image as $\xi_k(t_j)$ 
for all $j \in \{0,1,\cdots,M_{\tau}\}$. 
The training of the delay system is based on 
the gradient-based optimization using 
the Adam optimizer with a batch size of 100. 
The maximum number of epochs to train was set as 50 to ensure the convergence 
of the training process.
As a demonstration of the classifications at epoch 50, we show four examples of the softmax outputs, which represents the probability that the input image belongs to one of the 10-classes, in Figs. \ref{fig_dmr}(a)--(d). 

Figures \ref{fig_dmr}(e) and \ref{fig_dmr}(f) show 
the classification accuracy for various values of 
feedback strength $\beta$ and delay time $\tau$, where $T/\tau = 3$ 
is fixed. 
For this image dataset, the delay system with $\beta = 3.0$
exhibits relatively 
high classification performance.  
The best classification accuracy for this system is 97 $\%$. 
We emphasize that accurate classification is achieved 
with two training signals, $u_1(t), u_2(t)$, and minimal weight parameters, 
$\omega_l(t)$, and $b_l$, $l \in \{0,1,\cdots,9\}$, owing to the 
time-division multiplexing encoding method based on the delay structure, 
as discussed in Sec. \ref{sec_delay}.
This is in contrast to conventional neural networks, 
where more than hundreds or thousands of weight parameters need to be trained. 

We can see that classification accuracy improves
as delay time $\tau$ increases [Fig.\ref{fig_dmr}(f)]. 
As discussed in Sec. \ref{sec_delay}, 
the effective number of the network nodes, $M_{\tau}$, 
depends on $\tau$ in the delay system, i.e., $M_{\tau} \approx \tau/\Delta t$.
This suggests that 
larger-scale networks, i.e., systems with a longer delay, 
play an important role in achieving 
better classification for this large dataset. 

\section{Conclusion}
We discussed the applicability of optimally-controlled dynamical systems
to information processing. 
The dynamics-based processing provides insight into 
the mechanism of information processing based on deep network structures, 
and it can be easily implemented in physical systems. 
As a particular example, we introduced an optoelectronic delay system.
The delay system can be trained to perform nonlinear classification and image recognition with only a few control signals and classification weights 
based on the time-division multiplexing method.  
This feature of delay systems is an advantage to 
hardware implementation of the systems and 
is distinctively different from 
conventional neural networks, which require a large number of 
training parameters. 
The dynamics-based processing based on optimal control
can be applied to various physical systems to construct not only 
feedforward networks but also recurrent neural networks, including 
reservoir computing.
This provides a novel direction for physics-based computing.  
    
\begin{acknowledgements}
This work was supported, in part, by JSPS KAKENHI (Grant
No. 20H042655) and JST PRESTO (Grant No. JPMJPR19M4).
The authors thank Profs. Kazutaka Kanno and Atsushi Uchida for
valuable discussions on optoelectronic delay systems.  
\end{acknowledgements}

\appendix
\section{ Adjoint method for Eq. (\ref{eq1}) \label{app1}}
We derive the adjoint equations used to find an optimal 
control vector $\uu^*(t)$.
The first step is to incorporate the constraint of Eq. (\ref{eq1}), 
$\pdev{\rr}{t}{} - \FF(\rr,\uu) = {\bf 0}$, 
into loss function $J$ with the
Lagrangian multiplier (adjoint) vector $\pp_k(t) \in \RR^M$ as follows: 
\begin{align}
J_L = \sum_{k=1}^K
\left[
\Psi(\tm_k,\yy_k) + 
\int_0^T \pp_{k}^\tr
\left(
\FF_{k} 
- \dot{\rr}_{k}
\right)dt
\right], \label{app1-eq1}
\end{align}
where $\rr_{k} \in \RR^M$ is the state vector starting from the
initial state, $\rr_{k}(0) = \xx_k \in \RR^M$, and 
$\FF_{k} = \FF(\rr_{k},\uu(t))$.
Considering that the second term of Eq. (\ref{app1-eq1}) is rewritten as 
\begin{align}
 \int_0^T \pp_{k}^\tr
\left(
\FF_{k} 
- \dot{\rr}_{k}
\right)dt
&= 
- \pp_k^\tr(T)\rr_k(T)
+ \pp_k^\tr(0)\rr_k(0) \nonumber \\
&+ \int_0^T 
\left(
\pp_{k}^\tr 
\FF_{k} 
+
\dot{\pp}_{k}^\tr
\rr_{k}
\right)
dt,
\end{align}
we can obtain
\begin{align}
J_L &= \sum_{k=1}^K
\left[
\Psi(\tm_k,\yy_k)
- \pp_k^\tr(T)\rr_k(T)
+ \pp_k^\tr(0)\rr_k(0) 
\right] \nonumber \\
&+ \sum_{k=1}^K
\int_0^T 
\left(
\pp_{k}^\tr 
\FF_{k} 
+
\dot{\pp}_{k}^\tr
\rr_{k}
\right)
dt.
\label{app1-eq2}
\end{align}
Let $\delta \rr_k$ and $\delta J_L$ 
be the variations of $\rr_k$ and loss function $J_L$
in terms of the variation 
$\delta\uu$, respectively.
Then, variation $\delta J_L$ is computed as follows:
\begin{widetext}
\begin{align}
\delta J_L
&=
\sum_{k=1}^K
\left(
\left.
\del{\Psi}{\rr_k}{}
\right|_{t=T}
-
\pp_k^\tr(T)
\right) \delta\rr_k(T)
+
\sum_{k=1}^K
\int_0^T
\left[
\pp_k^\tr
\left(
\del{\FF_k}{\rr_k}{}\delta\rr_k
+
\del{\FF_k}{\uu}{}\delta\uu
\right)
+
\dot{\pp}_k^\tr\delta\rr_k
\right]
dt \nonumber \\
&=
\sum_{k=1}^K
\left(
\left.
\del{\Psi}{\rr_k}{}
\right|_{t=T}
-
\pp_k^\tr(T)
\right) \delta\rr_k(T)
+
\sum_{k=1}^K
\int_0^T
\left(
\pp_k^\tr
\del{\FF_k}{\rr_k}{}
+
\dot{\pp}_k^\tr
\right)\delta\rr_k
dt 
+\sum_{k=1}^K
\int_0^T
\pp_k^\tr
\del{\FF_k}{\uu}{}\delta\uu
dt, 
\label{app1-eq4}
\end{align}
\end{widetext}
where $\pdel{\Psi}{\rr_k}{} = \pdel{\Psi[\tm_k,\yy(\rr_k,\oomega)]}{\rr_k}{}$, 
and $\delta\rr_k(0) = 0$ is used in the aforementioned derivation 
because the initial state $\rr_k(0)$ is fixed as $\rr_k(0) = \xx_k$.
As the Lagrangian multiplier, $\pp_k$, can be set freely,
we select $\pp_k$ 
such that it satisfies the following equation:
\begin{align}
\pp_k^\tr(T) &= \left.
\del{\Psi_k}{\rr_k}{}
\right|_{t=T}, 
\mbox{\hspace{3mm} for $t = T$} \\
\dev{{\pp}_{k}^T}{t}{}
&= -\pp_{k}^\tr\del{\FF_{k}}{\rr_{k}}{}, 
\mbox{\hspace{3mm}for $0 < t < T$}. 
\label{app1-eq5} 
\end{align}
In this case, we obtain a simple form of $\delta J_L$ as follows:
$\delta J_L = \sum_k \int_0^T \pp_k^{\tr}\pdel{\FF_k}{\uu}{}\delta\uu dt$.
Accordingly, when the variation $\delta \uu$ is set as 
\begin{align}
\delta\uu = 
-\alpha\sum_{k=1}^K
\left(
\pp_k^{\tr}\del{\FF_k}{\uu}{} 
\right)^\tr,
\end{align}
with a positive constant $\alpha$, 
$\delta J_L = -1/\alpha\int_0^T\delta\uu^2 dt \le 0$ is satisfied. 
Thus, $J$ monotonically decreases when $\uu(t)$ is updated 
as $\uu(t) \rightarrow \uu(t) + \delta\uu$.

\section{Adjoint method for Eq. (\ref{delay-eq}) \label{app2}}
We consider the update variations, 
$\delta \uu(t)$, $\delta\oomega(t)$, and $\delta\bb$,
for the optimization of the delay system that obeys Eq. (\ref{delay-eq}).
In the same manner as that shown in Appendix \ref{app1}, we consider 
the augmented loss
function $J_L$ incorporating Eq. (\ref{delay-eq}) as follows:
\begin{align}
J_L = \sum_{k=1}^K
\Psi\left[
\tm_k,\yy_k
\left(\zz_k\right)
\right]
+
 \sum_{k=1}^K
\int_0^T\pp_{k}^\tr
\left(
\FF_{k}
-\dot{\rr}_{k}
\right)dt, \label{app2-eq1}
\end{align}
where 
$
\zz_k = \int_{T-\tau}^T \oomega\rr_kdt + \bb
$, 
$
\FF_{k} = \FF[\rr_{k}(t),\rr_{k}(t-\tau),\uu(t)]
$, 
and 
$\pp_{k}$ is the Lagrangian multiplier (adjoint vector).

Variation $\delta J$ in terms of the variation $\delta \uu$ is 
expressed as 
\begin{align}
\delta J_L &=
\sum_{k=1}^K 
\left(
\del{\Psi}{\zz_k}{}\int_{T-\tau}^T\oomega(t)\delta\rr_kdt
- \pp_k^{\tr}(T)\delta \rr_k(T)
\right) \nonumber \\
&+\sum_{k=1}^K 
\int_0^T
\pp_k^{\tr}
\left(
\del{\FF_k}{\rr_k}{}\delta\rr_k
+\del{\FF_k}{\rr_{k,\tau}}{}
\delta\rr_{k,\tau}
+ \del{\FF_k}{\uu}{}\delta\uu
\right)dt \nonumber \\
&+ 
\sum_{k=1}^K 
\int_0^T\dot{\pp}_k^{\tr}\delta\rr_k
dt,
\end{align}
where $\rr_{k,\tau} = \rr_k(t-\tau)$. 
In the aforementioned equation, considering 
$
\int_0^T\pp_k^{\tr}(t)\pdel{\FF_k}{\rr_{k,\tau}}{}\delta\rr_{k,\tau}dt
= 
\int_0^{T-\tau}\pp_k^{\tr}(t+\tau)\pdel{\FF_k(t+\tau)}{\rr_{k}}{}\delta \rr_kdt
$, 
variation $\delta J_L$ is rewritten as
\begin{widetext}
\begin{align}
\delta J_L 
&= -\sum_{k=1}^K\pp_k^{\tr}(T)\delta\rr_k(T) 
+ \sum_{k=1}^K
\int_{T-\tau}^T
\left(
\del{\Psi}{\zz_k}{}\oomega(t) + \pp_k^{\tr}(t)
\del{\FF_k}{\rr_k}{} + \dot{\pp}^{\tr}_k(t)
\right)\delta \rr_k dt \nonumber \\
&+\sum_{k=1}^K
\int_0^{T-\tau}
\left(
\pp_k^{\tr}(t)
\del{\FF_k}{\rr_k}{}
+
\pp_k^{\tr}(t+\tau)
\del{\FF_k}{\rr_k}{}(t+\tau)
+\dot{\pp}_k^{\tr}(t)
\right)
\delta\rr_k dt 
+
\sum_{k=1}^K
\int_0^T
\pp_k^{\tr}(t)
\del{\FF_k}{\uu}{} \delta\uu dt.
\end{align} 
\end{widetext}
When $\pp_k(t)$ is selected such that 
$\pp_k(T) = 0$ and
the following equations are satisfied, 
\begin{align}
\dev{\pp_{k}^\tr (t)}{t}{}
= 
-\del{\Psi}{\zz_{k}}{}\oomega(t)
-
\pp_{k}^{\tr}(t)
\dfrac{\partial \FF_{k}}{\partial \rr_{k}},
\end{align}
for $T-\tau \le t < T$, 
and 
\begin{align}
\dev{\pp_{k}^\tr (t)}{t}{}
=
-\pp^\tr (t)
\dfrac{\partial \FF_{k}}{\partial \rr_{k}}
-\pp_{k}^\tr (t+\tau)
\dfrac{\partial \FF_{k}(t+\tau)}{\partial \rr_{k}},
\end{align}
for $0 \le t < T-\tau$,
we can obtain a simple form of $\delta J_L$, 
as $\sum_{k=1}^K
\int_0^T
\pp_k^{\tr}(t)
\pdel{\FF_k}{\uu}{} \delta\uu dt.
$
When $\delta \uu$ is selected as
\begin{align} 
\delta \uu = 
-\alpha 
\sum_{k=1}^K
\left(
\pp_k^{\tr}(t)
\del{\FF_k}{\uu}{}
\right)^\tr,
\end{align} 
$\delta J_L = -1/\alpha \int \delta\uu^2 dt \le 0$ is always satisfied.

Then, we consider variations $\delta_{\oomega} J_L$ 
and $\delta_{\bb} J_L$ in terms of the variations of 
weights and bias parameters, respectively. 
In the same manner shown earlier, we obtain, 
$ \delta_{\oomega} J_L
= \sum_{k=1}^K
\pdel{\Psi}{\zz_k}{}\delta\zz_k
=
\sum_{k=1}^K
\pdel{\Psi}{\zz_k}{}
\int_{T-\tau}^T\delta\oomega\rr_k dt 
$
in terms of the weight variation $\delta\oomega$ 
and 
$ \delta_{\bb} J_L
= \sum_{k=1}^K
\pdel{\Psi}{\zz_k}{}\delta\bb
$
in terms of the weight variation $\delta\bb$. 
Clearly, when $\delta\oomega$ and $\delta \bb$ 
are set as follows:
\begin{align}
\delta \oomega 
= -\alpha_{\omega}\sum_{k=1}^K
\left(\del{\Psi}{\zz_k}{}
\right)^{\tr}
\rr_k^\tr, 
\end{align}
and 
\begin{align}
\delta \bb 
= -\alpha_{b}\sum_{k=1}^K
\left(\del{\Psi}{\zz_k}{}
\right)^{\tr},
\end{align}
with small positive constants, $\alpha_\omega$ and $\alpha_b$, 
$\delta_{\oomega} J_L \le 0$ and $\delta_{\bb} J_L \le 0$. 

\section{Update equations for $u_1$, $u_2$, $\omega_l$, and $b_l$ 
in the optoelectronic delay system \label{sec_app2-2}}
In this study, we set the loss function as 
$J = -1/K\sum_{k=1}^{K}\sum_{l=0}^{L-1} t_{l,k}\log y_{l,k}$, 
where
$y_{l,k} = e^{z_{l,k}}/\sum_{l=0}^{L-1} e^{z_{l,k}}$,  
and 
$z_{l,k} = \int_{T-\tau}^T\omega_{l}(t)\xi_k(t) dt + b_l$. 
In this case, the adjoint equations for $\pp_k = (p_{\xi,k},p_{\eta,k})^{\tr}$ 
are given as follows for $T-\tau \le t < T$,
\begin{align}
\dev{{p}_{\xi,k}}{t}{} &= -\dfrac{1}{K}\sum_{l=0}^{L-1}(y_{l,k}-t_{l,k})\omega_{l} + g
p_{\xi,k} -g_H p_{\eta,k}, \\
\dev{p_{\eta,k}}{t}{} &= g_L p_{\xi,k},
\end{align}
and for $T < t \le T-\tau$, 
\begin{align}
\dev{{p}_{\xi,k}}{t}{} &= 
gp_{\xi,k} -g_H p_{\eta,k} +\tilde{\beta}u_{1}(t+\tau)\sin\delta_k(t+\tau)p_{\xi,k}(t+\tau),\\
\dev{p_{\eta,k}}{t}{} &= g_L p_{\xi,k}, 
\end{align}
where $g = 1/\tau_H + 1/\tau_L$, $g_H = 1/\tau_H$, $g_L = 1/\tau_L$,  
$\tilde{\beta} = \beta/\tau_L$, 
and $\delta_k(t) = 2[u_{1}(t)\xi_k(t-\tau) +u_{2}(t)]$.
Then, the update variables of control signal vectors are given as
follows:
\begin{align}
\delta u_1(t)
&= \alpha_u\tilde{\beta}\sum_k^Kp_{\xi,k}\sin\delta_k(t)\xi_{k}(t-\tau), \\
\delta u_2(t)
&=
\alpha_u
\tilde{\beta}
\sum_k^Kp_{\xi,k}\sin\delta_k(t),
\end{align}
and the update variables of weights and bias parameters
are 
\begin{align}
\delta\omega_l
&=
-\dfrac{\alpha_{\omega}}{K}
\sum_k^K(y_{l,k}-t_{l,k})\xi_k, \\
\delta b_l
&=
-\dfrac{\alpha_b}{K}
\sum_k^K(y_{l,k}-t_{l,k}).
\end{align}

\end{document}